\documentclass[preprint,12pt]{elsarticle}

\usepackage{graphicx} 
\usepackage{multirow}
\usepackage{subcaption}
\usepackage{hyperref}
\usepackage{xcolor}
\usepackage[a4paper, total={6in, 8in}]{geometry}
\usepackage{rotating} 

\begin{document}

\begin{frontmatter}

\title{Reducing the gap between general purpose data and aerial images in concentrated solar power plants}

\author[1,2]{M.A. Pérez-Cutiño} 
\author[1,2]{J. Valverde} 
\author[3]{J. Capitán} 
\author[2]{J.M. Díaz-Báñez} 

\affiliation[1]{organization={Virtualmechanics S.L.},
            city={Seville},
            country={Spain}}

\affiliation[2]{organization={Department of Applied Mathematics II, University of Seville},
            city={Seville},
            country={Spain}}

\affiliation[3]{organization={Multi-robot \& Control Systems group, University of Seville},
            city={Seville},
            country={Spain}}

\begin{abstract}

In the context of Concentrated Solar Power (CSP) plants, aerial images captured by drones present a unique set of challenges. Unlike urban or natural landscapes commonly found in existing datasets, solar fields contain highly reflective surfaces, and domain-specific elements that are uncommon in traditional computer vision benchmarks. As a result, machine learning models trained on generic datasets struggle to generalize to this setting without extensive retraining and large volumes of annotated data. However, collecting and labeling such data is costly and time-consuming, making it impractical for rapid deployment in industrial applications.  

To address this issue, we propose a novel approach: the creation of AerialCSP, a virtual dataset that simulates aerial imagery of CSP plants. By generating synthetic data that closely mimic real-world conditions, our objective is to facilitate pretraining of models before deployment, significantly reducing the need for extensive manual labeling.
Our main contributions are threefold: (1) we introduce AerialCSP, a high-quality synthetic dataset for aerial inspection of CSP plants, providing annotated data for object detection and image segmentation; (2) we benchmark multiple models on AerialCSP, establishing a baseline for CSP-related vision tasks; and (3) we demonstrate that pretraining on AerialCSP significantly improves real-world fault detection, particularly for rare and small defects, reducing the need for extensive manual labeling.
AerialCSP is made publicly available at \url{https://mpcutino.github.io/aerialcsp/}.

\end{abstract}



\end{frontmatter}


\newpage

\section{Introduction}

In recent years, deep learning has revolutionized computer vision, enabling machines to interpret complex visual data with remarkable accuracy. Many state-of-the-art models are first trained on large-scale, general-purpose datasets before being adapted to specific applications. This paradigm provides a solid foundation for feature learning but often falls short when models are transferred to specialized domains, where data distribution shifts and task-specific nuances emerge \cite{planamente2021da4event,doig2023improved}. The challenge is particularly pronounced in robotics and aerial vision, where models must operate in environments significantly different from those found in standard datasets \cite{nagananda2021benchmarking}. However, pretraining models on domain-specific datasets that closely resemble the target application yields superior performance and significantly reduces the annotation requirements when transferring knowledge to similar contexts.

\begin{figure}
    \centering
    \includegraphics[width=0.95\columnwidth]{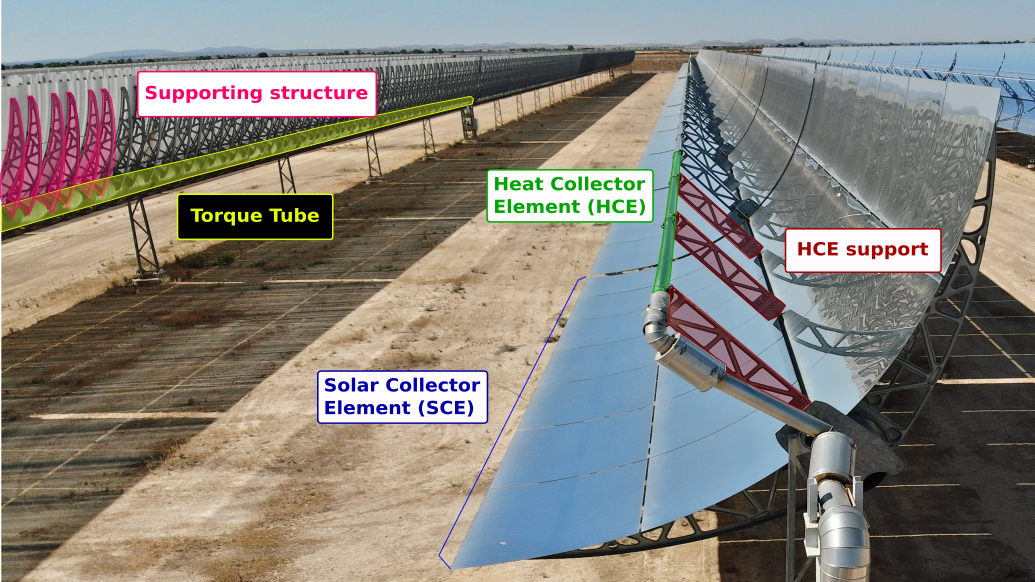}
    \caption{Elements in parabolic-trough CSP plants. The names of the elements are surrounded with bounding boxes, using different colors to identify them. Each name matches a class in AerialCSP dataset.}
    \label{fig:aerialcsp_elements}
\end{figure}

Concentrated Solar Power (CSP) is a promising renewable energy technology capable of meeting future energy demands sustainably. It offers the advantage of dispatchable electricity generation based on the incorporation of large-scale thermal energy storage in a cost-effective manner. The most common CSP technology is Parabolic Trough (PT) solar field~\cite{alami2023concentrating}. PT plants utilize a field of solar-tracking reflectors, parabolic mirrors, known as Solar Collector Elements (SCE), to continuously direct concentrated sunlight onto an absorber tube or Heat Collector Element (HCE). This concentrated solar radiation is converted into thermal energy by increasing the enthalpy of a Heat Transfer Fluid (HTF) running along the HCE, which is subsequently employed in a Rankine thermodynamic cycle to produce electricity or directly used as heat for industrial process applications. Some important elements in PT systems are depicted in Figure \ref{fig:aerialcsp_elements}.

Technical studies \cite{mcnamara2023estimating} highlight the critical role of automation in both the development and operational phases of CSP plants. Increasing automation enhances operational efficiency, reduces maintenance costs, and improves energy output, ultimately lowering the Levelized Cost of Electricity (LCOE) and increasing the economic viability of CSP technology. A key enabler of this automation is the integration of Unmanned Aerial Vehicles (UAVs), which provide a flexible, scalable, and highly efficient solution for monitoring and maintenance. Unlike ground-based or human-based inspection methods, UAVs offer uninterrupted operation without interfering with plant activities, and capturing high-resolution aerial data in minimal time. The modular design of PT solar fields further allows for the deployment of UAV fleets, significantly accelerating the characterization of large-scale CSP facilities.  

Various strategies have been proposed for fault detection in commercial CSP plants using UAV-based inspections. In \cite{perez2024meta}, a methodology is introduced to assess breakages in the glass envelopes of HCEs in the solar field, leveraging computer vision and deep learning classifiers. Similarly, \cite{perez2023detecting} explores the extraction of thermal information from aerial images to evaluate the condition of HCE glass envelopes.
Several studies have focused on mirror and absorber tube shape and alignment issues. In \cite{prahl2013airborne}, UAVs were used to measure slope deviations in mirrors, employing SCE corners and coded markers for accurate detection. This approach was later refined in \cite{prahl2017absorber} to detect HCE displacements relative to their ideal positions. A complementary method was proposed in \cite{kesseli2023combined}, where deep learning algorithms were combined with traditional computer vision techniques for large-scale optical analysis in parabolic-trough solar plants. Their approach enabled automated SCE identification and image segmentation, improving the robustness of mirror characterization.
Other UAV-based approaches have targeted thermal diagnostics in CSP plants. In \cite{lamghari2019innovative}, UAV imagery was combined with data from a car-mounted capture system to detect broken thermal insulation in ball joints and assess soiling levels on SCEs. 
A more recent study in~\cite{perez2024measuring} focuses on the evaluation of ball joint assemblies to detect HTF leakages and structural imperfections.
In the context of heliostat field inspection using aerial imagery, \cite{tian2022toward} proposes a flight path optimization method aimed at enhancing contrast in polarized images to improve edge detection and crack identification on mirror surfaces. 
For a broader review of UAV applications in CSP plants, 
we refer the reader to the recent survey conducted by \cite{milidonis2023unmanned}.

While existing research has primarily focused on fault detection in solar power systems, many of these approaches rely on custom datasets specifically tailored to the problem at hand. This reliance on manually collected and annotated data makes such methods labor-intensive and time-consuming. One fundamental limitation is that most machine learning models are initially trained on general-purpose datasets that lack domain-specific features representative of CSP environments. Although techniques such as few-shot 
\cite{tang2023privacy}
and zero-shot \cite{bansal2018zero} learning can partially mitigate data requirements, a more comprehensive strategy is needed to bridge the gap between generic datasets and the unique visual characteristics of CSP plants. To address this, we propose a novel, domain-adaptive approach that leverages a virtual dataset for object detection in CSP facilities.
Our objective is not to curate a narrowly tailored data set to optimize performance on a single inspection task, but rather to develop a general purpose synthetic data set that enables effective pre-training of deep learning models. These pretrained models can then be fine-tuned for a wide range of downstream applications, such as tracking angle estimation, broken component detection, mirror misalignment identification, HCE localization in thermal imagery, among others. Despite the diversity of these tasks, they share a common spatial and visual context, that of identifying key elements within the solar field. By capturing this shared structure, our dataset reduces the need for extensive annotated data across multiple inspection pipelines, promoting reusability and scalability.

\subsection{Literature review}

Creating custom datasets for real-world applications is often essential when addressing new problems using supervised learning techniques. These datasets are commonly used to pretrain models, which immediately enables knowledge transfer to related real-world tasks. When real-world datasets include diverse and representative object classes, they are especially valuable for transfer learning, with ImageNet \cite{russakovsky2015imagenet} and COCO \cite{lin2014microsoft} serving as foundational examples in general-purpose object detection. This strategy has also been adopted in domains such as railway infrastructure \cite{qiu2024whu}, urban scenes \cite{cordts2015cityscapes}, and photovoltaic systems \cite{wang2024pvf}, where task-specific datasets have driven notable progress. However, to the best of our knowledge, in the context of PT plants there is currently no publicly available dataset tailored for object detection tasks. This gap is largely due to the relative novelty of CSP technology and the reluctance of plant operators to release aerial imagery, often due to privacy or proprietary concerns. In this setting, a synthetic dataset like AerialCSP provides a compelling alternative, offering a way to train and share robust models while protecting sensitive operational data, while also reducing the gap between generic vision datasets and the specific needs of PT plant maintenance and monitoring.

The idea of creating virtual datasets 
has been widely applied in the existing literature.
For instance, \cite{zhao2025construction} utilizes Unity3D to generate a virtual dataset simulating aerial imagery for maritime search and rescue missions. These images are entirely created through simulation, which often requires intricate scene design or compromises in realism, particularly in background rendering. Similarly, \cite{broda2023towards} employs a Blender-based simulation pipeline to produce synthetic aerial views of heliostat fields for airborne monitoring. Although effective for several computer vision tasks, their method struggles to distinguish between reflective mirror surfaces and background textures. Platforms such as AirSim \cite{shah2018airsim} have also been employed to simulate urban and forest environments, providing high-fidelity synthetic data for autonomous navigation tasks.
Alternative virtual dataset creation strategies include copy-paste techniques, in which segmented objects are composited onto varied background images with random transformations such as rotation, translation, and scaling \cite{xu2022data, perez2024measuring}. Another interesting method is domain randomization, which introduces variation in the appearance of the modeled object, particularly useful when shape information is more critical than texture, as demonstrated in \cite{tobin2017domain, barisic2022sim2air}. 
More recently, virtual data generation using generative deep learning models has gained momentum, as these models improve in producing highly diverse and photorealistic images \cite{cai2020piigan,horvath2022metgan}. Nevertheless, in industrial applications where the environment is well defined and can be accurately replicated using simulation tools, physics-based simulation remains the preferred strategy.

\subsection{Contribution}

To the best of our knowledge, this is the first work to directly address the gap between general-purpose datasets and a domain-specific dataset tailored for aerial imagery in Parabolic Trough solar fields. Unlike prior approaches for synthetic data generation that either rely on arbitrary randomization or demand highly detailed scene modeling, our method introduces a novel background generation technique that balances realism with efficiency. This allows for better domain adaptation without compromising simulation speed or requiring exhaustive scene configuration. Furthermore, our dataset is publicly released to foster further research in the field of autonomous inspection and monitoring of Parabolic Trough facilities.

We summarize our contributions as follows:
\begin{itemize}
    \item \textbf{High-quality synthetic dataset for CSP Inspection.} We introduce and release AerialCSP, a virtual dataset designed specifically for aerial inspection of Parabolic Trough plants. Our methodology preserves the original plant environment images while integrating simulated elements. 
    To the best of our knowledge, this is the first publicly available dataset containing aerial images from Parabolic Trough solar fields.
    \item \textbf{Benchmarking and analysis.} We conduct extensive benchmarking of YOLOv11 \cite{yolo11_ultralytics} models on AerialCSP, evaluating their performance on both object detection and image segmentation. Our results establish a baseline for models trained on AerialCSP.
    \item 
    \textbf{Case study: the impact of AerialCSP on real-world CSP fault detection.}
    We conduct a case study using 800 real aerial images to assess the effectiveness of pretraining with AerialCSP for fault detection tasks in CSP plants. Our experiments show that models pretrained on AerialCSP require fewer annotated samples to achieve high performance and outperform models trained on general-purpose data, particularly in detecting rare and small-scale defects such as broken HCEs and damaged mirrors.
\end{itemize}

In particular, our case study underscores the advantages of using AerialCSP as a proxy dataset to facilitate downstream tasks that would otherwise require more than twice the amount of annotated data. The remainder of the paper is structured as follows: Section~\ref{sec:virtualds} presents the framework for constructing AerialCSP, and Section~\ref{sec:experiments} benchmarks and evaluates its effectiveness for knowledge transfer in real-world CSP applications. Finally, Section~\ref{sec:conclusions} discusses the limitations and broader implications of this work.

\section{Virtual Dataset}
\label{sec:virtualds}

\begin{figure}[t]
    \centering
    \includegraphics[width=.98\linewidth]{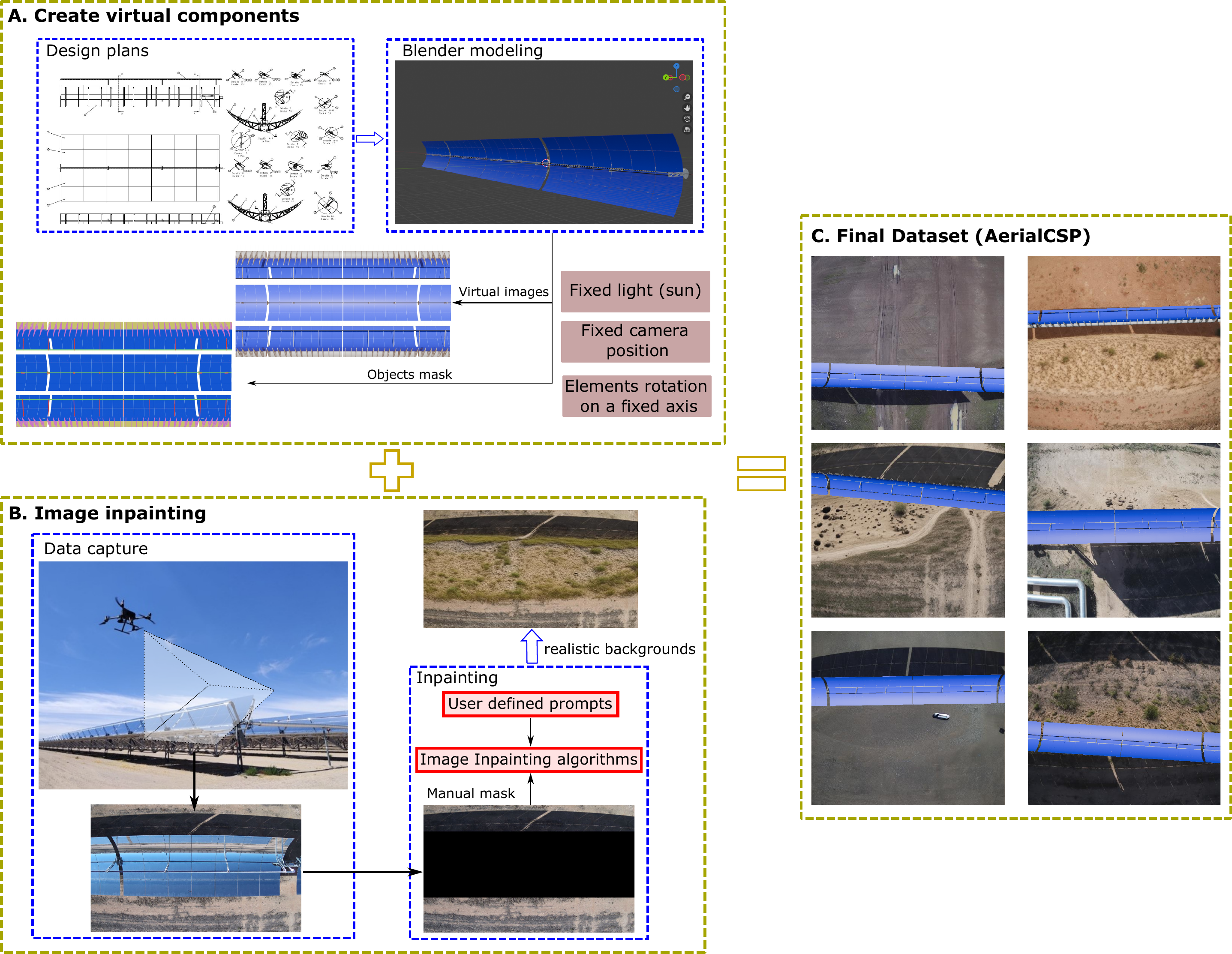}
    \caption{Overview of the proposed methodology for creating AerialCSP.}
    \label{fig:proposal_overview}
\end{figure}

The process of creating the virtual dataset is outlined in Figure \ref{fig:proposal_overview}. We structure this process into several key modules, which are described in detail in the following sections.

\subsection{3D Modeling}

To accurately replicate the structural components of a Solar Collector Element, we begin by importing its design plans into Computer Aided Design (CAD) software. These models are then exported as STL files and subsequently imported into Blender, where we enhance their realism using Blender’s scripting capabilities.

A fixed light source is integrated into the Blender environment and positioned perpendicular to the X–Y plane to emulate direct sunlight. A virtual camera is placed between the light source and the SCE to replicate the aerial viewpoint of a UAV. To simulate real-world CSP plant operation, the SCE is rotated incrementally around the Y-axis to mimic solar tracking. The tracking step size is set to one degree, and for each tracking position, two images are generated:
\begin{itemize}
    \item A photorealistic rendering of the SCE, providing a visually accurate representation of its components.
    \item A segmentation mask, where each pixel value corresponds to a distinct object class within the image.
\end{itemize}

\begin{figure}[t]
    \centering
    \begin{subfigure}{.49\columnwidth}
    \centering
\includegraphics[width=\columnwidth,trim={5cm 18cm 30cm 14.5cm},clip]{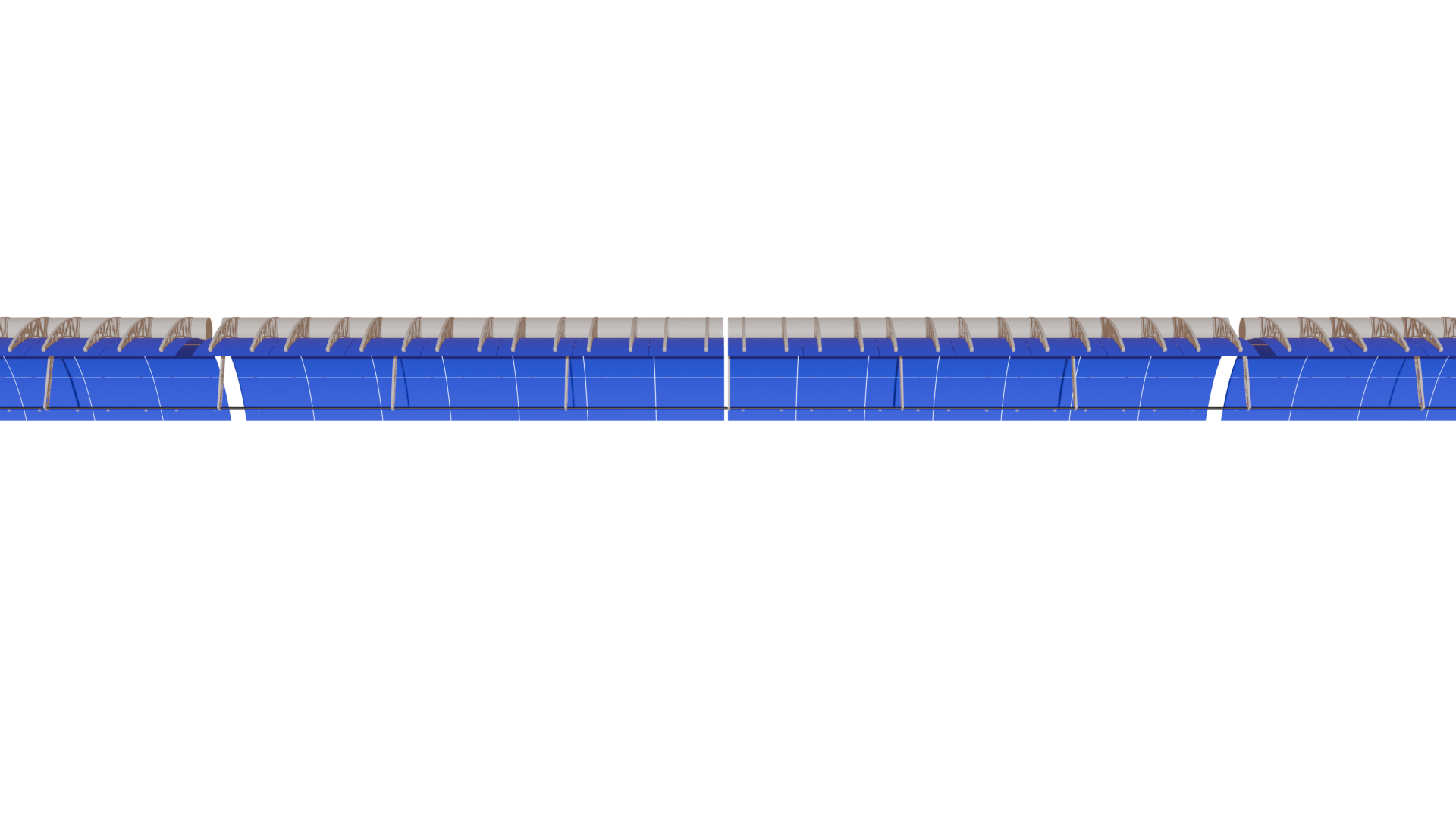}
    \end{subfigure}
    \hfill
    \begin{subfigure}{.49\columnwidth}
        \includegraphics[width=\columnwidth,trim={5cm 18cm 30cm 14.5cm},clip]{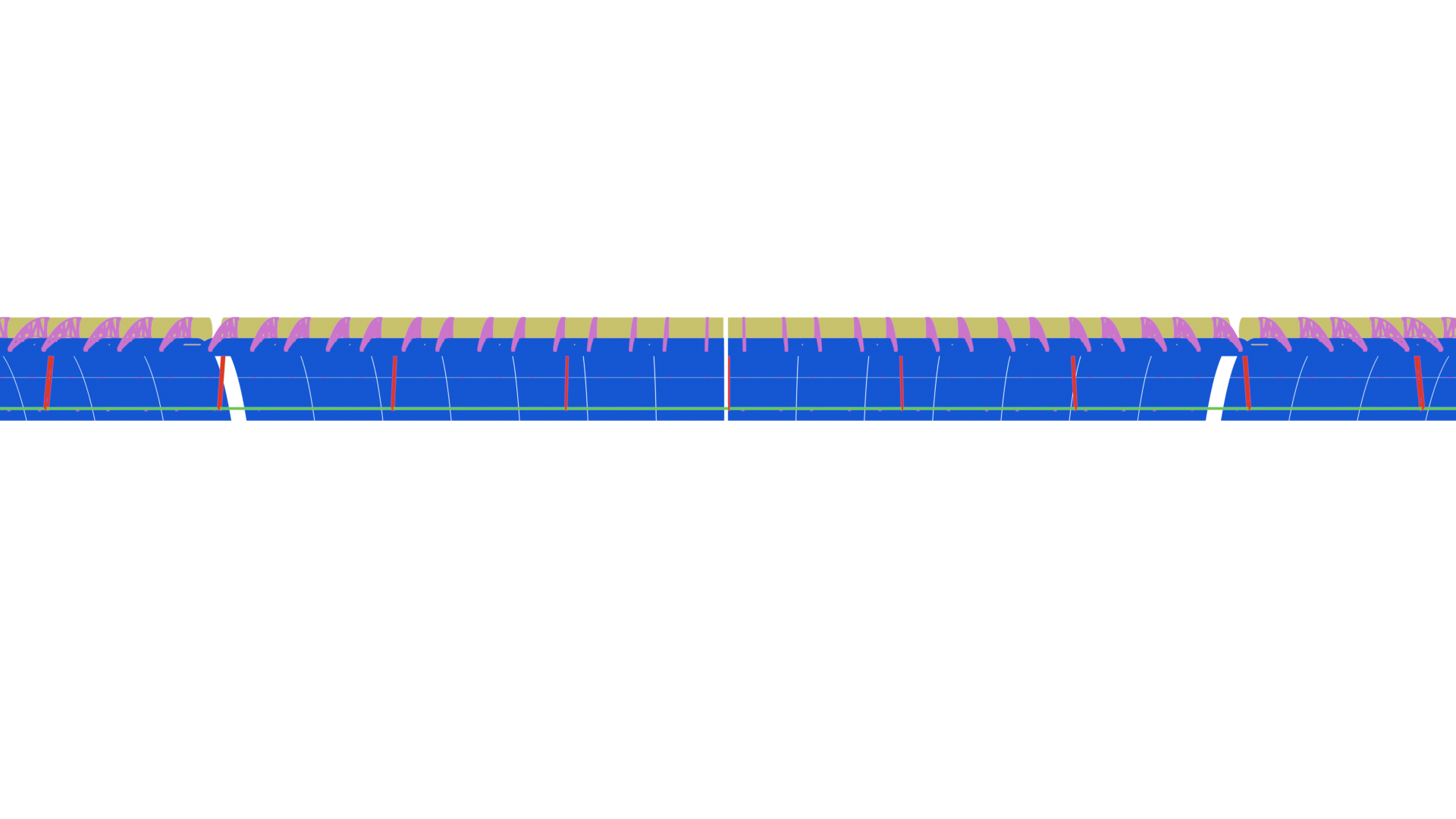}
    \end{subfigure}
    \caption{Resulting images from 3D modeling in Blender. (left) Photorealistic rendering of the components of the solar collector, (right) segmentation mask where each object class is assigned a different color.}
    \label{fig:blender_output_example}
\end{figure}

The photorealistic image and the segmentation mask are both captured for the same scene. Prior to capturing, objects in the scene are assigned surface materials according to the intended output. For the photorealistic render, materials are designed to reflect sunlight, cast realistic shadows, and closely mimic real-world conditions. In contrast, the segmentation render uses flat, solid colors to represent each object class, enabling precise extraction and labeling of target elements within the scene.
An example of these image pairs is shown in Figure \ref{fig:blender_output_example}. The segmentation mask assigns unique colors to each object class: SCEs or \texttt{mirrors} (blue), \texttt{HCE support} (red), \texttt{HCE} (green), \texttt{torque tube} (yellow), and \texttt{supporting structure} (pink). These labeled masks serve as ground truth annotations for downstream machine learning tasks.

\subsection{Background Inpainting}

\begin{figure}[t]
    \centering
    \begin{subfigure}{0.95\textwidth}
    \centering
        \includegraphics[width=0.32\textwidth,height=3.3cm]{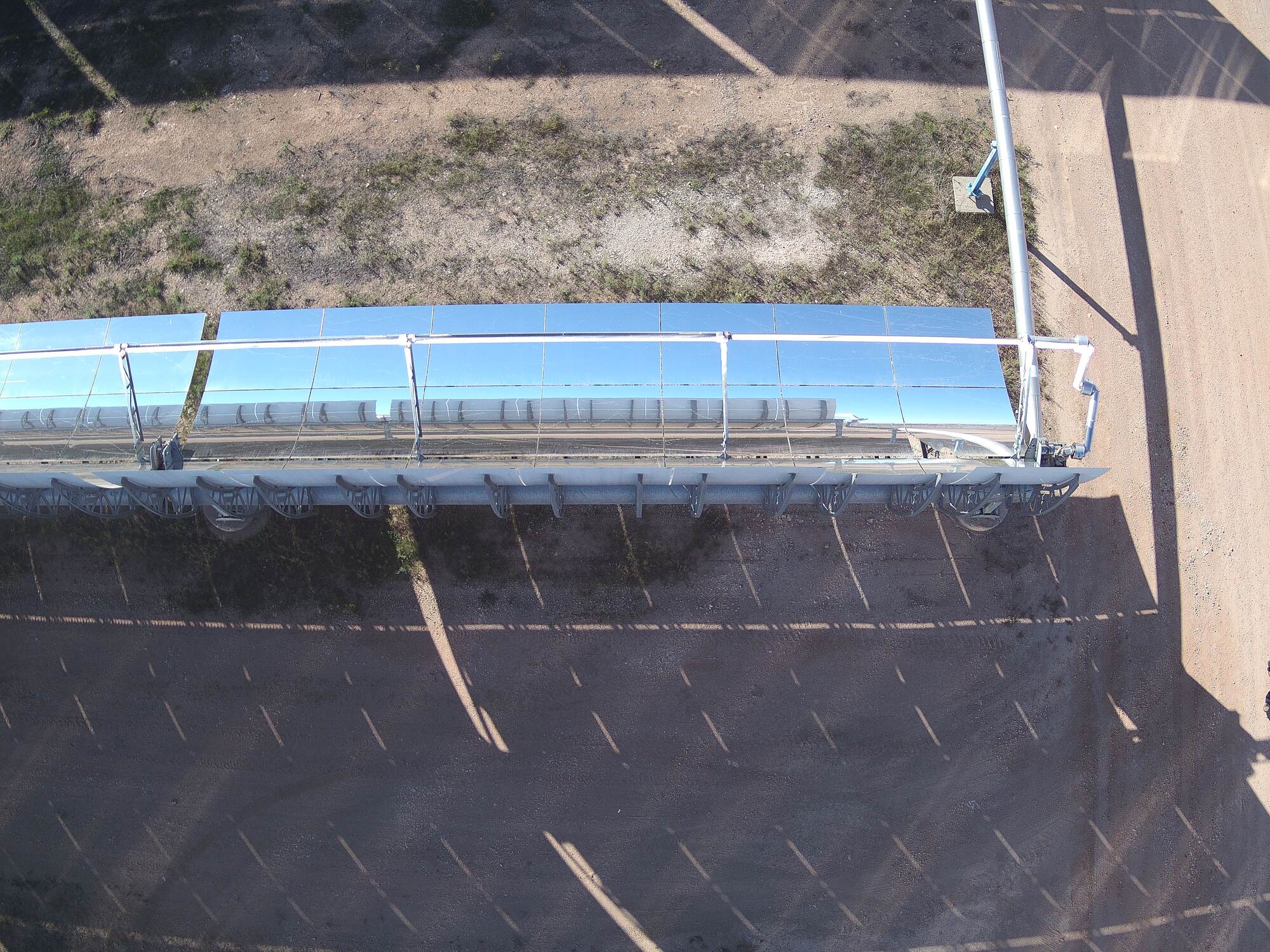}
        \hfill
        \includegraphics[width=0.32\textwidth,height=3.3cm]{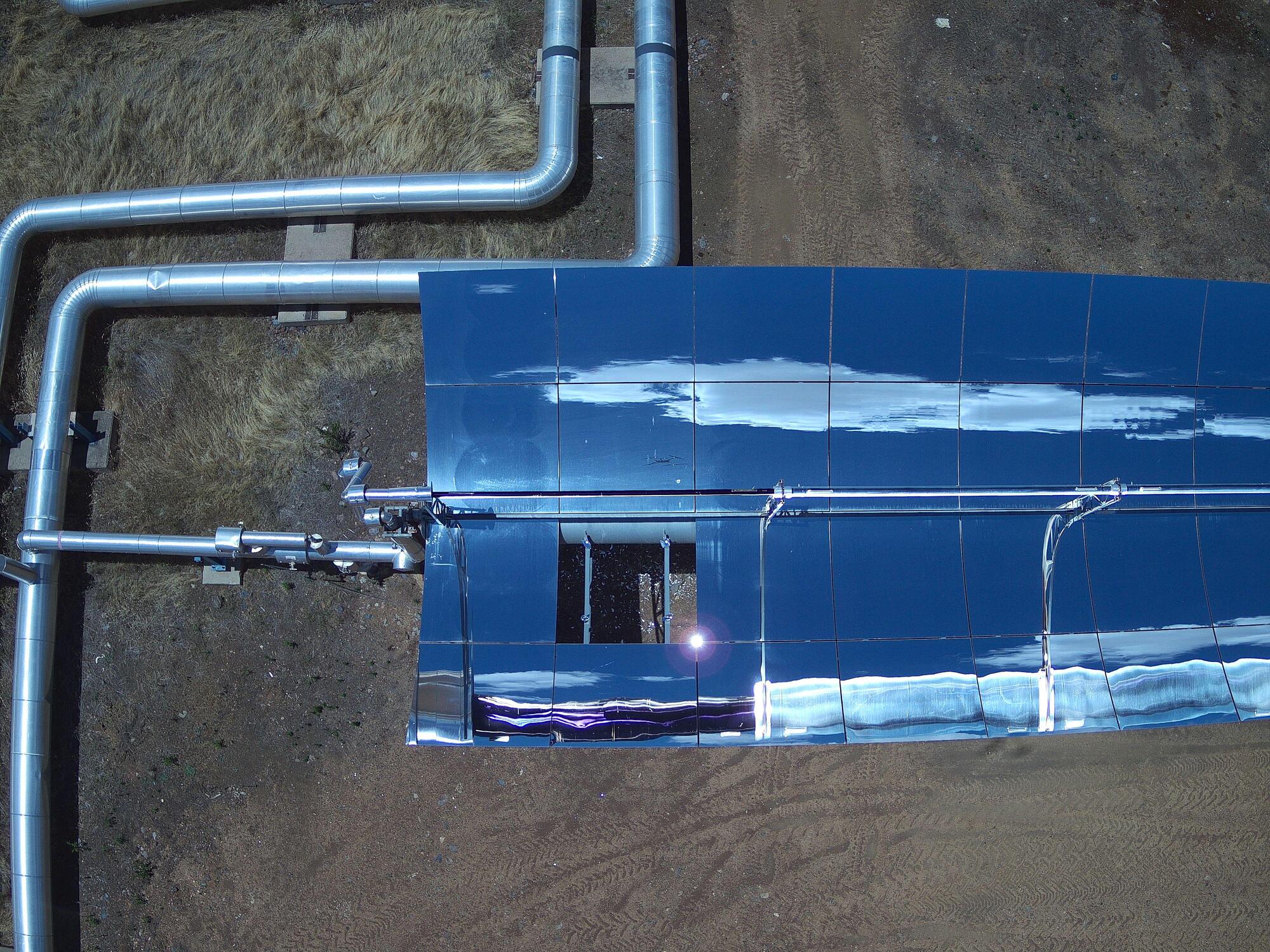}
        \hfill
        \includegraphics[width=0.32\textwidth,height=3.3cm]{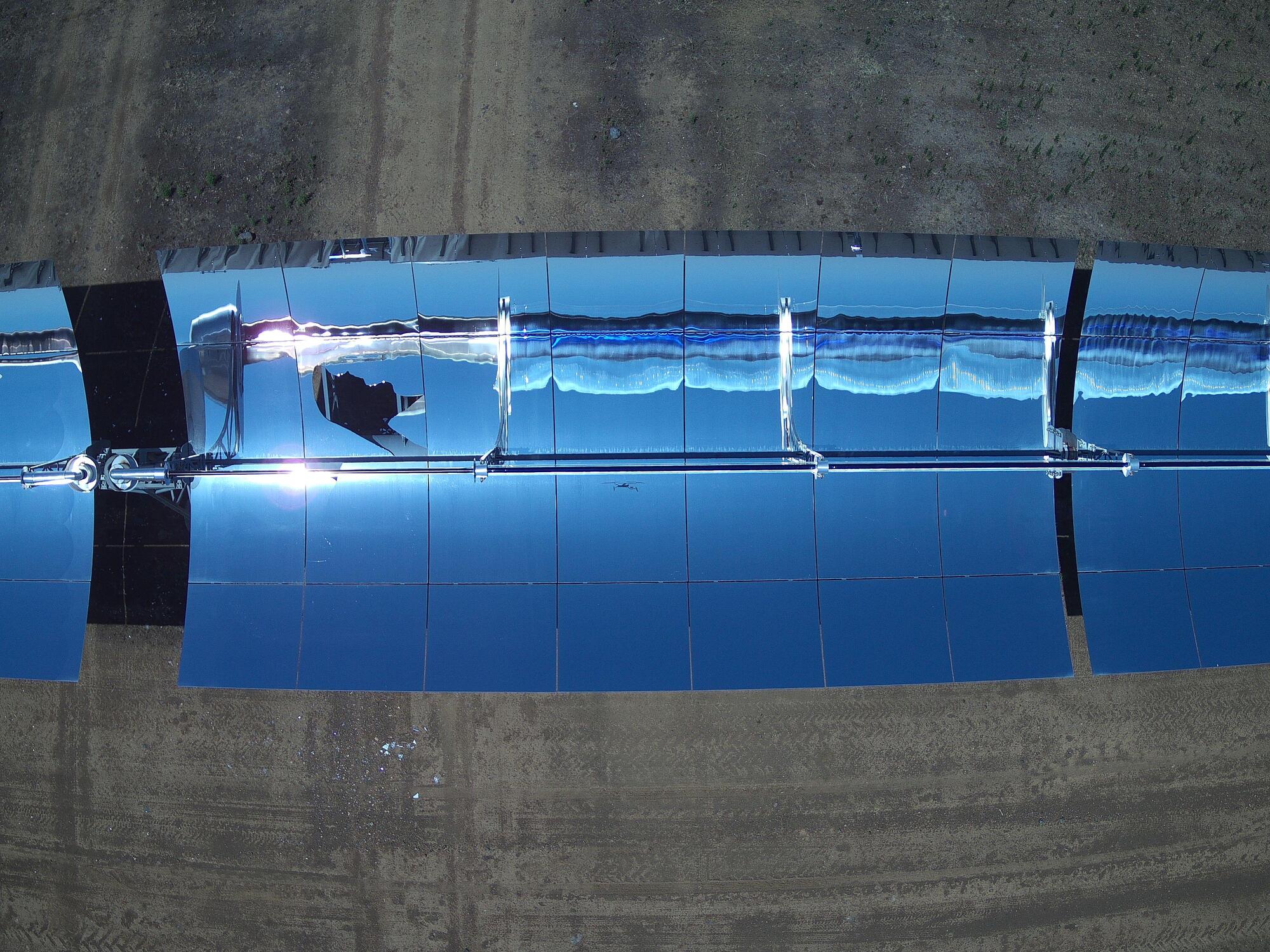}
        \caption{}
    \end{subfigure}

    \begin{subfigure}{0.95\textwidth}
    \centering
        \includegraphics[width=0.32\textwidth,height=3.3cm]{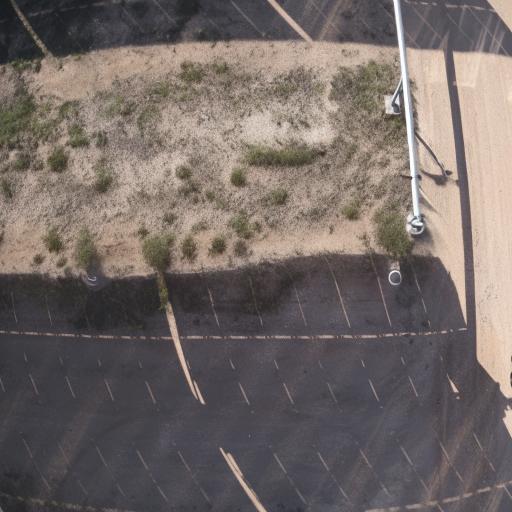}
        \hfill
        \includegraphics[width=0.32\textwidth,height=3.3cm]{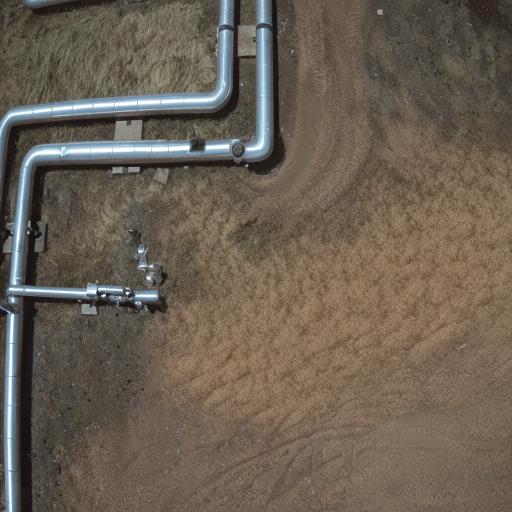}
        \hfill
        \includegraphics[width=0.32\textwidth,height=3.3cm]{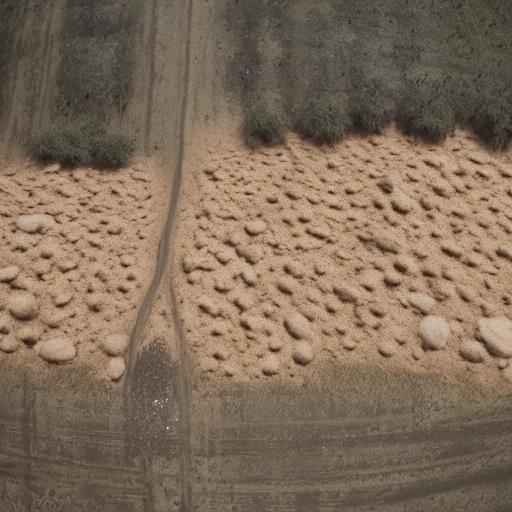}
        \caption{}
    \end{subfigure}

    \begin{subfigure}{0.95\textwidth}
    \centering
        \includegraphics[width=0.32\textwidth,height=3.3cm]{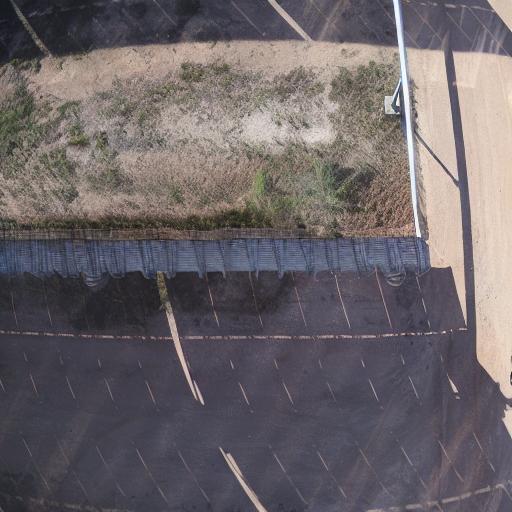}
        \hfill
        \includegraphics[width=0.32\textwidth,height=3.3cm]{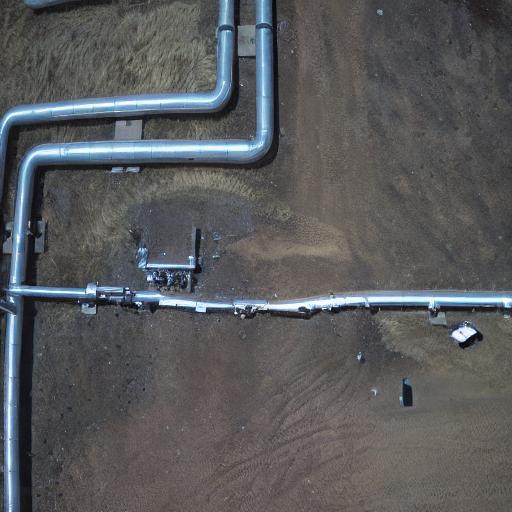}
        \hfill
        \includegraphics[width=0.32\textwidth,height=3.3cm]{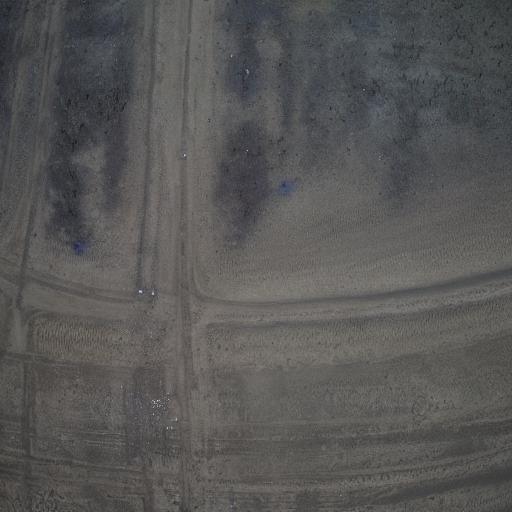}
        \caption{}
    \end{subfigure}
    
    \caption{Background inpainting in aerial images of CSP plants, removing 3D-modeled objects such as SCEs, HCEs, HCEs supports, torque tubes, and supporting structures. (a) Real images captured from a UAV, (b) results obtained with Kandinsky algorithm, (c) results obtained with Stable Diffusion algorithm.}
    \label{fig:inpainting_examples}
\end{figure}

To enhance the realism of the dataset, we create synthetic backgrounds from natural CSP plant environments. We achieve this using image inpainting techniques. 
In general, image inpainting algorithms reconstruct missing or damaged portions of an image by filling in the gaps with plausible content.
We exploit this property by removing objects from the original image and relying on a deep learning model to reconstruct the missing areas. This process enables us to produce realistic, unobstructed backgrounds that can later be used to insert simulated objects and build a virtual dataset.

Our approach uses a set of 800 real-world images captured with an UAV deployed in several Parabolic Trough solar fields. 
The UAV used for data collection was the DJI Matrice 300 RTK, operating at altitudes between 20 and 23 meters (typical for combined thermal and visual inspections). The onboard camera was oriented vertically downward, with the gimbal set to a -90° angle. For data preprocessing, a rectangular region within each image containing unwanted objects, such as SCEs and HCEs, was identified. This region served as a mask input to the inpainting algorithm, specifying the area of the image to be reconstructed. This allows us to obtain a realistic background of the solar field, free from structural components, which can then be replaced with virtual counterparts. It is important to note that we do not need to manually label every element in the image to produce the mask; instead, it suffices to annotate the broader region containing all relevant structures, as illustrated in Figure \ref{fig:proposal_overview}.

We leverage Kandinsky \cite{razzhigaev-etal-2023-kandinsky} and Stable Diffusion \cite{rombach2022high}, two state-of-the-art inpainting models, to reconstruct the background. These models take as input the original image along with the mask indicating the regions to be replaced. Additionally, a text prompt can be provided to guide the appearance of the reconstructed content. 
Throughout our experiments, we tested multiple prompt configurations to optimize image quality. We found that Kandinsky was more user-friendly, yielding acceptable realism with minimal prompt tuning, but Stable Diffusion produced superior results when effective prompts were carefully crafted. However, we observed that the best-performing prompt for one image did not necessarily generalize well to others. Therefore, we selected a fixed prompt that we identified as good for a randomly selected subset of the initial images. 
Consequently, the final dataset exhibits a range of image quality levels using both Kandinsky and Stable Diffusion algorithms. Nevertheless, our experiments confirm that models trained on AerialCSP demonstrate substantial performance gains compared to models trained on generic datasets. Figure \ref{fig:inpainting_examples} illustrates the diversity of inpainted backgrounds.

\subsection{Final Dataset}
\label{subsec:final_ds}

The final AerialCSP dataset is created by compositing the rendered Blender images onto the inpainted backgrounds. To introduce variability, we apply random scaling and rotations to the synthetic SCE images before pasting them onto the backgrounds. These transformations are also applied to the corresponding segmentation masks to maintain annotation consistency. AerialCSP is designed to support two primary computer vision tasks:
\begin{itemize}
    \item \textbf{Object detection:} Each image is accompanied by a YOLO-formatted annotation file containing bounding box information. The annotation format follows \texttt{<class> <x\_center> <y\_center> <width> <height>}, where all coordinates are normalized (0 to 1), and class indices are zero-indexed.
    \item \textbf{Instance segmentation:} Segmentation masks are stored in a format where each line corresponds to an object and follows \texttt{<class-index> <x$_1$> <y$_1$> <x$_2$> <y$_2$> ... <x$_n$> <y$_n$>}, where (x, y) coordinates define the polygonal mask of the object and are normalized within [0,1].
\end{itemize}

Images are generated by selecting each simulated object from the 3D modeling stage and applying random transformations (translations, rotations, and rescales) before pasting it onto a randomly selected background produced by the inpainting module. This process is carried out separately for each of the two inpainting algorithms used. Since there is one simulated object for each solar tracking angle, a total of 181 unique objects were placed onto background images, and this procedure was repeated 100 times (50 per inpainting algorithm) to generate a diverse dataset. After generation, 42 low-quality images were manually removed due to objects being positioned too close to the image boundaries. In total, AerialCSP contains 18,058 labeled images.

The AerialCSP dataset is publicly available at \url{https://mpcutino.github.io/aerialcsp/}, 
facilitating research in aerial robotics and automated monitoring of CSP plants. To ensure fair comparisons, we provide a standardized dataset split into training, validation, and test sets, following a 75\%-15\%-10\% distribution. 

\section{Experiments}
\label{sec:experiments}

We conduct two sets of experiments to demonstrate the advantages of AerialCSP. First, we benchmark our virtual dataset for both bounding box detection and image segmentation tasks. Then, we manually label a set of 800 real images for the task of failure detection in CSP plants and evaluate the impact of pretraining on AerialCSP, demonstrating its effectiveness in real-world applications.

To assess model performance, we use the following standard metrics:
\begin{itemize}
    \item Precision (P): Measures the accuracy of detected objects, indicating the proportion of correct detections among all predictions.
    \item Recall (R): Evaluates the model’s ability to identify all instances of objects in the images.
    \item mAP50: Mean Average Precision calculated at an Intersection over Union (IoU) threshold of 0.50, measuring detection accuracy for relatively easy cases.
    \item mAP50-95: The average mAP computed over multiple IoU thresholds, ranging from 0.50 to 0.95, providing a more comprehensive evaluation across varying detection difficulties.
\end{itemize}
Intersection over Union (IoU) quantifies the overlap between predicted and ground-truth objects, serving as the basis for these metrics.   

For our experiments, we evaluate different variants of YOLOv11 \cite{yolo11_ultralytics}, a state-of-the-art object detection framework. YOLO models offer architectures of varying sizes, making them suitable for different computational environments:
\begin{itemize}
    \item YOLOv11n: A lightweight model optimized for resource-constrained environments, such as UAV-based platforms.
    \item YOLOv11m: A mid-size model offering a balance between accuracy and computational efficiency.
    \item YOLOv11x: The largest model, delivering high precision but requiring substantial computational resources.
\end{itemize}
These models are also used as backbones for image segmentation tasks, where additional layers are appended to generate pixel-wise object masks. In these cases, for presenting our results, we append \emph{-seg} to the model name.

\subsection{Benchmarking the Dataset}

To benchmark AerialCSP, we train three different YOLOv11 models for 200 epochs, evaluating their performance on both bounding box detection and image segmentation tasks. We report the results obtained for the model with a higher score in the validation set during training, taking into account the dataset split described in Section \ref{subsec:final_ds}. The objective of this experiment is to provide a comparison point for future works.

Table \ref{tab:detection_results_aerialCSP} presents benchmark results for both bounding box detection and image segmentation on the AerialCSP test set, which consists of 1,807 images. Performance is reported per model and per class.

For the task of bounding box detection, all three models achieve exceptionally high scores, surpassing 90\% mAP50 for any class in the dataset, and 61\% for the mAP50-95. These results highlight the effectiveness of AerialCSP for object detection in CSP environments. The class with the lowest performance is \texttt{HCE support}, likely due to its small size and frequent occlusion when the collector is rotated during solar tracking.

In contrast, image segmentation results, shown in the right portion of Table \ref{tab:detection_results_aerialCSP}, exhibit a performance drop. While mAP50 remains close to 55\%, the mAP50-95 does not exceed 40\%, with the exception of the \texttt{mirror} class, which is arguably the easiest to detect due to its size and consistent appearance. Although this might seem like a significant reduction compared to bounding box detection, it is important to note that image segmentation is inherently a more complex task.
We attribute the lower performance primarily to the challenges of detecting small objects, which become more prominent when the Solar Collector Element is rotated. This is particularly evident in the segmentation results for the \texttt{HCE support}, \texttt{Supporting structure}, and \texttt{Torque tube} classes. Despite this, our goal with this benchmark is to establish a reference point for future research and facilitate domain adaptation before applying models to real-world CSP tasks.

\begin{sidewaystable}
    \centering
    \begin{tabular}{|c|c|c|c|c|c||c|c|c|c|c|c|}
    \hline
      \multirow{2}{*}{Models} & \multirow{2}{*}{Class}  &  \multicolumn{8}{c|}{AerialCSP Task} \\ \cline{3-10}
      & & \multicolumn{4}{c||}{Bounding Box Detection} & \multicolumn{4}{c|}{Image Segmentation} \\ \cline{3-10}
      & & P & R & mAP50 & mAP50-95 & P & R & mAP50 & mAP50-95 \\
      \hline
      yolo11n (-seg)    & \texttt{Mirrors} & \textbf{100} &  99.9 & 99.5 & \textbf{99.2} & 81.9 & 77.7 & 84.5 & 65.2 \\
      yolo11m (-seg)   & \texttt{Mirrors} & 99.9 & \textbf{100} & 99.5 & 89.5 & 82.6 & 83.4 & 88.0 & 68.4 \\
      yolo11x (-seg)  & \texttt{Mirrors} & 99.8 & \textbf{100} & 99.5 & 89.5 & \textbf{83.1} & \textbf{85.3} & \textbf{88.7} & \textbf{69.9} \\
      \hline
      yolo11n (-seg)   & \texttt{HCE} & \textbf{91.8} & 93.0 & 95.7 & 82.4 & 89.2 & 82.4 & 86.3 & 39.1 \\
      yolo11m (-seg)   & \texttt{HCE} & 91.0 & \textbf{93.2} & \textbf{96.2} & 84.8 & \textbf{96.6} & 88.1 & 91.6 & 37.1 \\
      yolo11x (-seg)   & \texttt{HCE} & 91.5 & \textbf{93.2} & \textbf{96.2} & \textbf{85.5} & 95.5 & \textbf{88.3} & \textbf{91.7} & \textbf{40.0} \\
      \hline
      yolo11n (-seg)   & \texttt{HCE support} & 95.9 & 76.2 & 90.2 & 61.4 & \textbf{72.2} & 45.6 & \textbf{55.9} & 25.6 \\
      yolo11m (-seg)   & \texttt{HCE support} & \textbf{96.9} & \textbf{78.4} & \textbf{92.0} & 66.7 & 67.2 & \textbf{49.9} & 55.6 & \textbf{25.7} \\
      yolo11x (-seg)   & \texttt{HCE support} & 96.8 & 77.4 & 91.5 & \textbf{69.8} & 65.1 & 47.1 & 54.3 & 25.1 \\
      \hline
      yolo11n (-seg)   & \texttt{Supporting structure} & 94.6 & 79.7 & 91.8 & 66.6 & \textbf{78.5} & 54.5 & 65.6 & 27.6 \\
      yolo11m (-seg)   & \texttt{Supporting structure} & 96.6 & \textbf{82.8} & \textbf{93.9} & 72.6 & 75.6 & \textbf{55.9} & 66.2 & 28.8 \\
      yolo11x (-seg)   & \texttt{Supporting structure} & \textbf{96.8} & 82.6 & 93.8 & \textbf{74.8} & 75.3 & 55.6 & \textbf{66.5} & \textbf{28.9} \\
      \hline
      yolo11n (-seg)   & \texttt{Torque tube} & \textbf{99.2} & 99.2 & 99.5 & 85.8 & 66.6 & 62.1 & 61.0 & 22.6 \\
      yolo11m (-seg)   & \texttt{Torque tube} & 97.0 & \textbf{99.6} & 99.5 & 86.5 & \textbf{68.8} & 64.7 & 64.8 & 25.8 \\
      yolo11x (-seg)   & \texttt{Torque tube} & 96.5 & 99.5 & 99.5 & \textbf{87.2} & 68.5 & \textbf{65.6} & \textbf{65.4} & \textbf{26.9} \\
      \hline
    \end{tabular}
    \caption{Object detection results using different YOLOv11 models in AerialCSP. Results from bounding box algorithms and image segmentation are reported separately. The suffix -seg indicates that the segmentation version of the model was used for the image segmentation task. We highlight in bold the highest value obtained per metric and class.
    }
    \label{tab:detection_results_aerialCSP}
\end{sidewaystable}

\subsection{Case study: transfer learning to fault detection}

To demonstrate the advantages of pretraining models on AerialCSP, we conduct an additional experiment using real-world data. We deploy a UAV across multiple CSP plants and manually annotate 800 images for the bounding box detection task. The dataset includes the following four classes: \texttt{mirrors}, \texttt{HCEs}, \texttt{broken HCEs}, and \texttt{broken mirrors}. While mirrors and HCEs are already present in AerialCSP, broken HCEs and broken mirrors introduce novel elements. Detecting these faults is critical for CSP operators, as early identification enables timely replacements, minimizing energy production losses.
We report results for YOLO models pretrained on AerialCSP, as well as models pretrained on a generic dataset such as COCO \cite{lin2014microsoft}. Since both models undergo a pretraining phase, we use the prefix \textit{pretrained} for models trained on AerialCSP, and \textit{not-pretrained} for those trained on COCO.

\textbf{Experimental setup.} From the 800 labeled images, we set aside 400 images as a test set and design five individual training experiments using subsets of the remaining 400 images. Each experiment progressively increases the training set size using 10, 50, 100, 200 and 400 images. This approach helps determine the minimum number of labeled images required to effectively adapt our models to the target task. During training, the data is augmented using random flips, cropping, and mosaic augmentation techniques.

\begin{figure}[t]
    \centering
    \begin{subfigure}{.49\columnwidth}
    \centering
        \includegraphics[width=\columnwidth,clip]{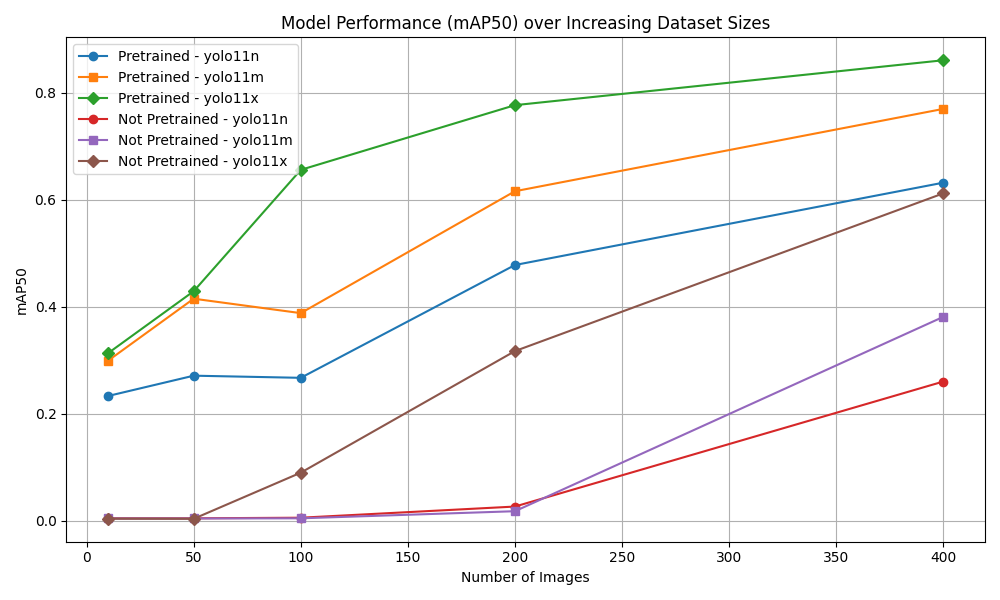}
        \caption{}
    \end{subfigure}
    \hfill
    \begin{subfigure}{.49\columnwidth}
        \includegraphics[width=\columnwidth]{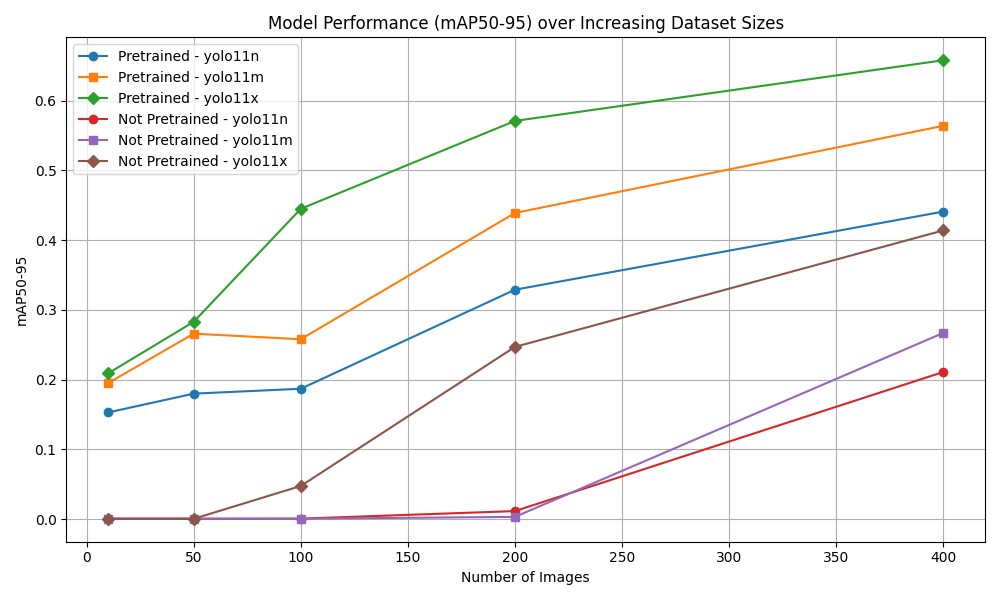}
        \caption{}
    \end{subfigure}
    \caption{Models performance in the task of bounding box detection when using real-world data. x-axis identify the number of images in the training set, while y-axis shows the perfomance of the models when using (a) the mAP50 metric and (b) the mAP50-95 metric. Pretrained prefix is used for models pretrained on AerialCSP. Metrics are averaged across all classes.}
    \label{fig:model_perfomances_real_data}
\end{figure}

\textbf{Results and analysis.} Figure \ref{fig:model_perfomances_real_data} illustrates the results across different training set sizes. The x-axis represents the number of training images, while the y-axis shows the model’s performance on the test set.
The benefits of pretraining on AerialCSP are evident:
\begin{enumerate}
    \item Models pretrained on AerialCSP consistently outperform their non-pretrained counterparts.
    \item Even the smallest pretrained model achieves better performance than the largest non-pretrained model. This is especially relevant for resource-constrained environments, such as aerial platforms.
    \item As the number of labeled images increases, the gap between pretrained and non-pretrained models narrows, suggesting that sufficient training data can eventually compensate for the lack of pretraining.
\end{enumerate}

\begin{table}
    \centering
    \begin{tabular}{|c|c|c|c|c|}
    \hline
       \multirow{2}{*}{Models} & \multicolumn{2}{c|}{Not pretrained} & \multicolumn{2}{c|}{Pretrained} \\ \cline{2-5}
       & mAP50 & mAP50-95 & mAP50 & mAP50-95 \\
      \hline
      yolo11n & 26.0 & 21.1 & \textbf{63.2} & \textbf{44.1} \\
      yolo11m & 38.1 & 26.7 & \textbf{77.0} & \textbf{56.4} \\
      yolo11x & 61.2 & 41.4 & \textbf{86.1} & \textbf{65.8} \\
      \hline
    \end{tabular}
    \caption{Object detection results for the task of object detection using different YOLOv11 models trained with 400 real images from thermosolar plants. Metrics are averaged across all classes.}
    \label{tab:realds400_detection_results}
\end{table}

\begin{table}
    \centering
    \begin{tabular}{|c|c|c|c|c|}
    \hline
       \multirow{2}{*}{Models} & \multicolumn{2}{c|}{Not pretrained} & \multicolumn{2}{c|}{Pretrained} \\ \cline{2-5}
       & b-HCE & b-Mirrors & b-HCE & b-Mirrors \\
      \hline
      yolo11n & 0.00 & 6.90 & \textbf{43.7} & \textbf{30.2} \\
      yolo11m & 3.99 & 23.6 & \textbf{77.0} & \textbf{56.4} \\
      yolo11x & 35.3 & 37.7 & \textbf{82.0} & \textbf{70.7} \\
      \hline
    \end{tabular}
    \caption{Object detection results for the mAP50 metric in the task of bounding box detection using different YOLOv11 models trained with 400 real images from thermosolar plants. These results only include the classes \texttt{broken HCE} (b-HCE) and \texttt{broken mirrors} (b-Mirrors).}
    \label{tab:realds400_detection_results_broken_things}
\end{table}

\begin{figure*}
    \centering
    \begin{subfigure}{0.95\textwidth}
    \centering
        \includegraphics[width=0.32\textwidth,height=3.5cm]{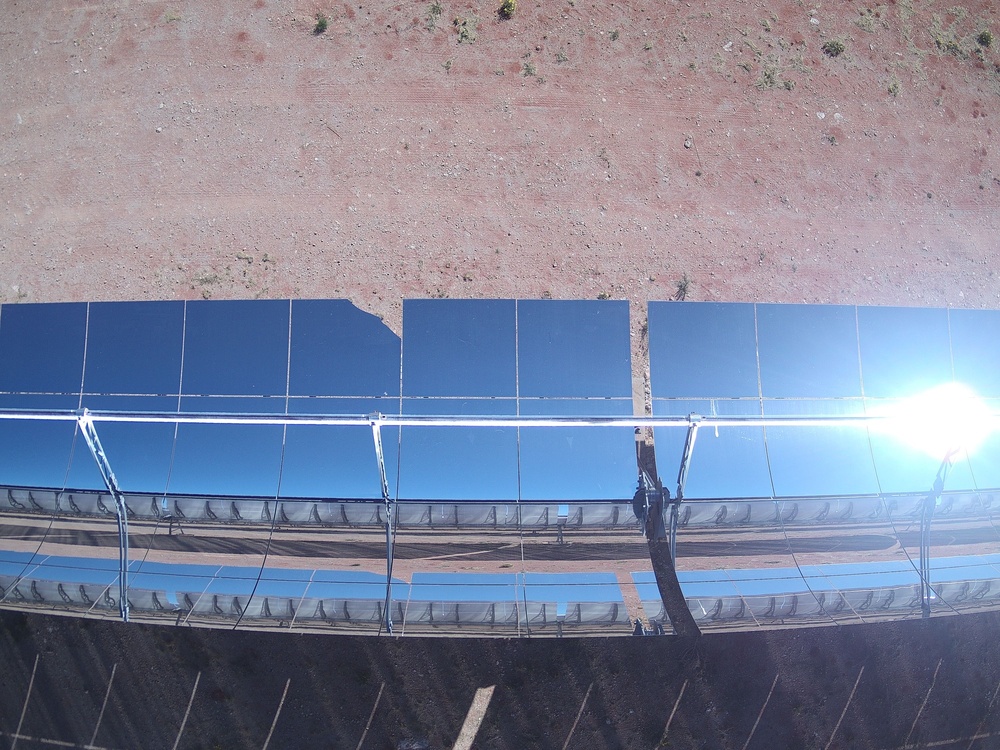}
        \hfill
        \includegraphics[width=0.32\textwidth,height=3.5cm]{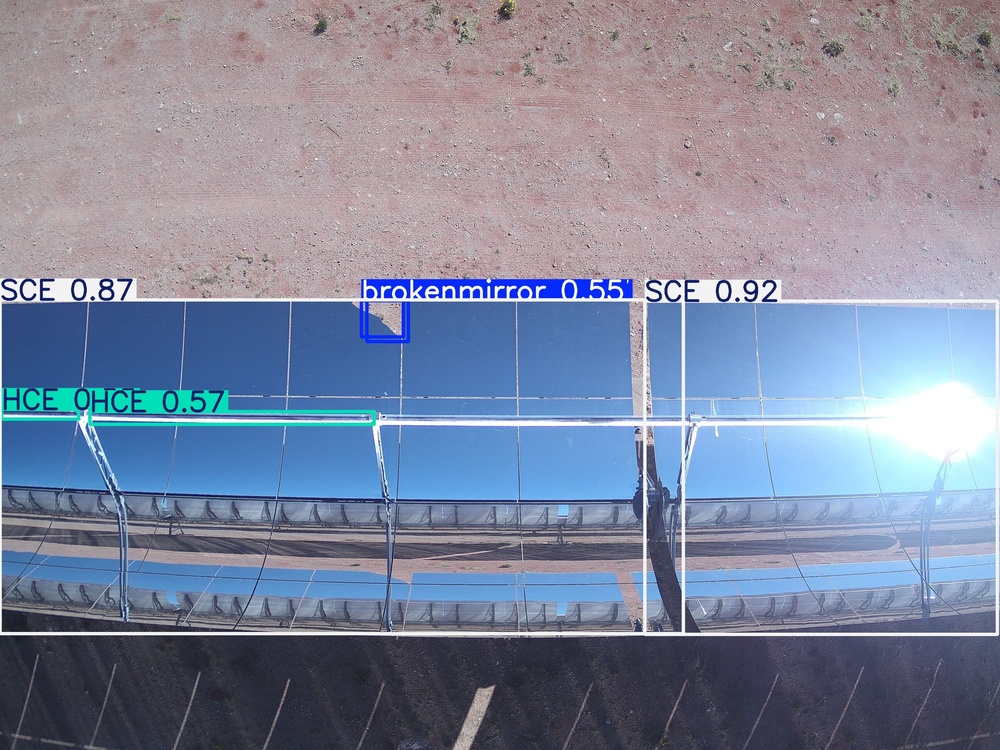}
        \hfill
        \includegraphics[width=0.32\textwidth,height=3.5cm]{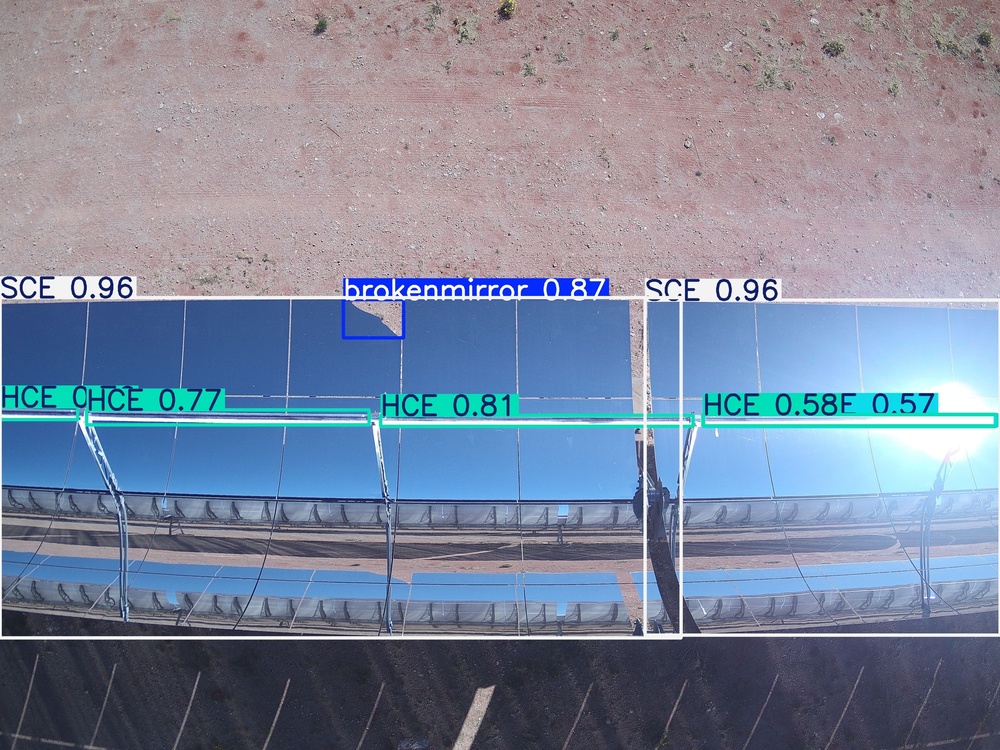}
        \caption{}
    \end{subfigure}

    \begin{subfigure}{0.95\textwidth}
    \centering
        \includegraphics[width=0.32\textwidth,height=3.5cm]{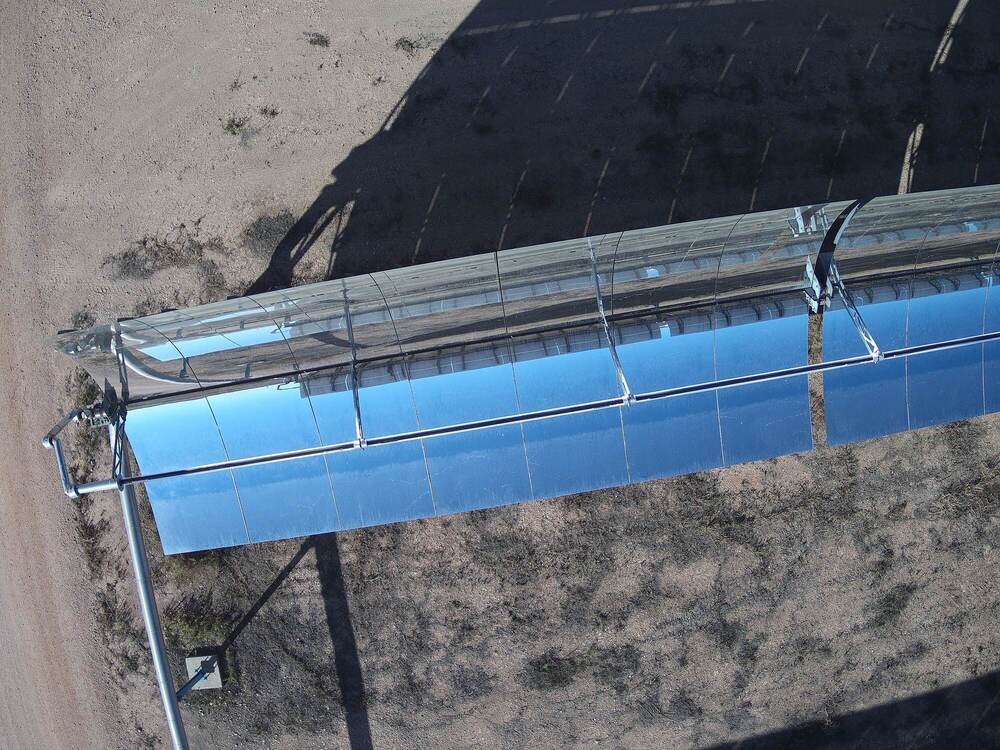}
        \hfill
        \includegraphics[width=0.32\textwidth,height=3.5cm]{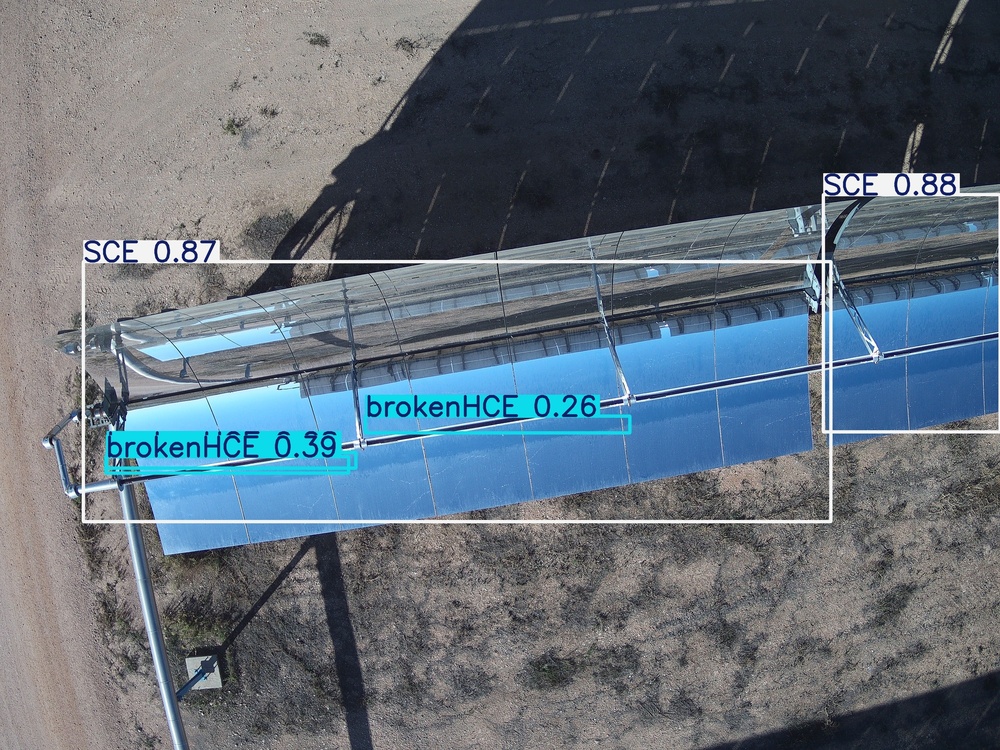}
        \hfill
        \includegraphics[width=0.32\textwidth,height=3.5cm]{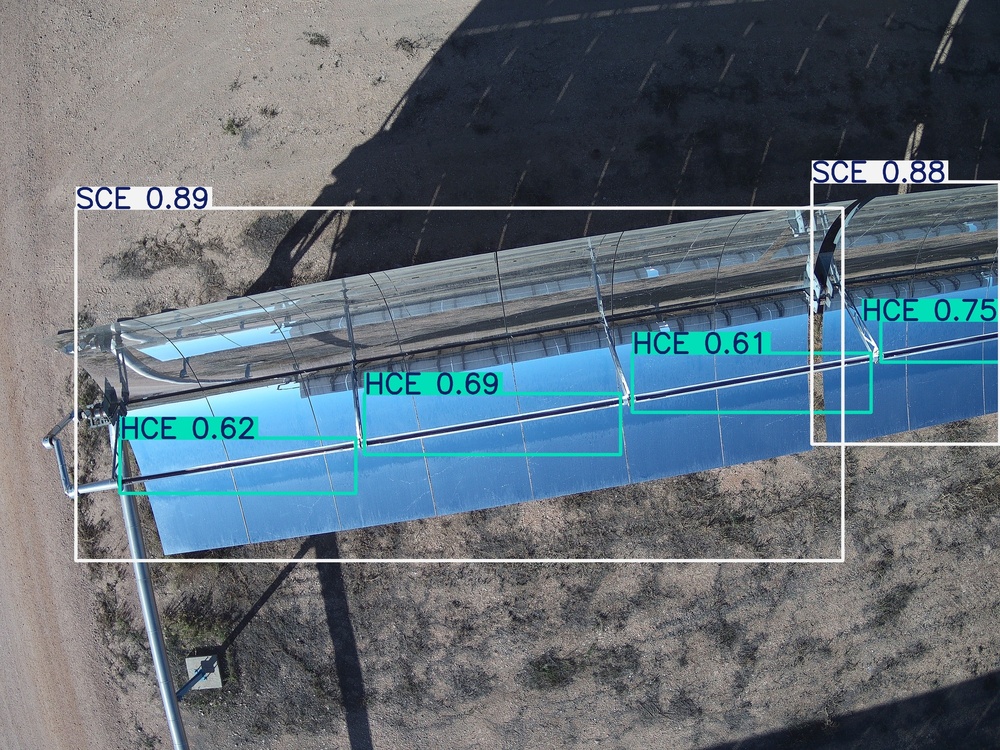}
        \caption{}
    \end{subfigure}

    \begin{subfigure}{0.95\textwidth}
    \centering
        \includegraphics[width=0.32\textwidth,height=3.5cm]{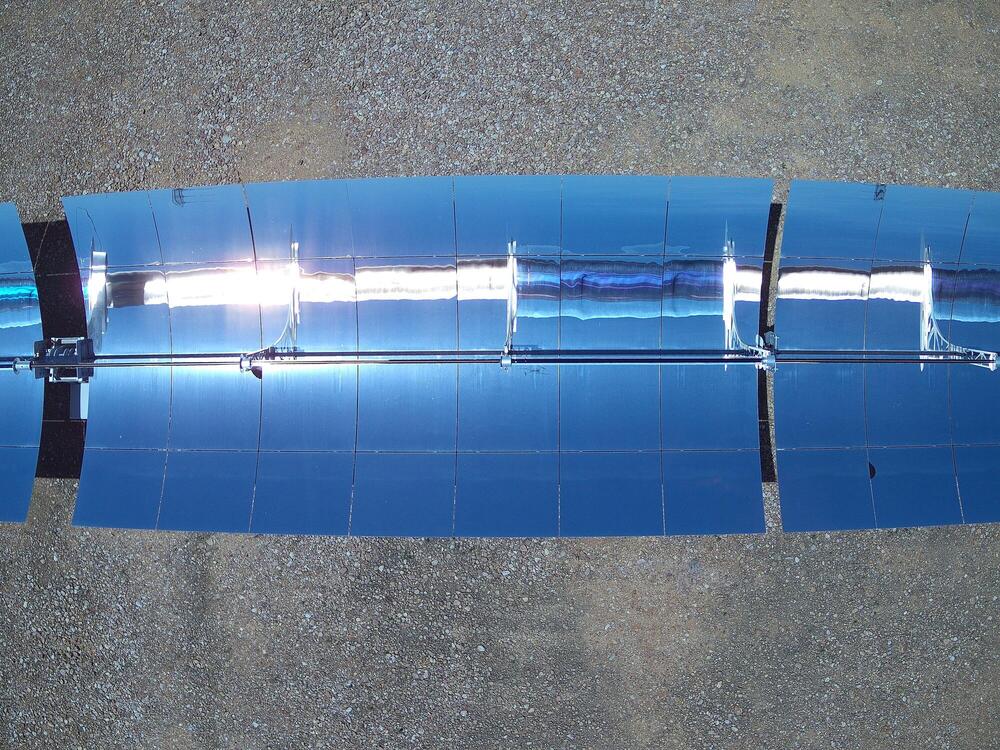}
        \hfill
        \includegraphics[width=0.32\textwidth,height=3.5cm]{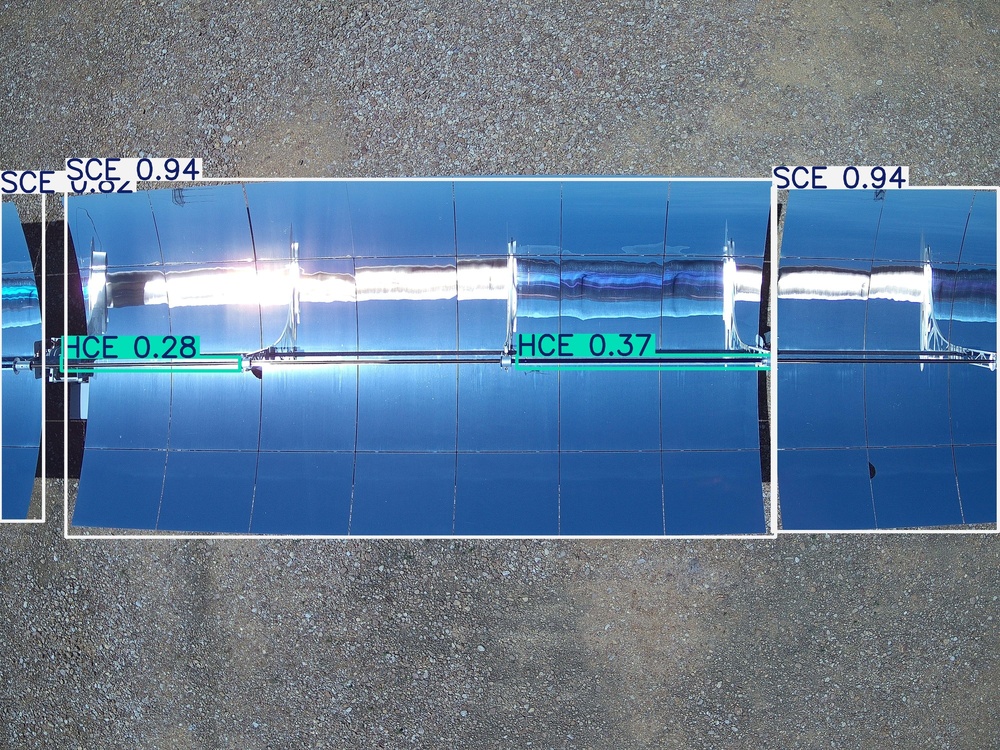}
        \hfill
        \includegraphics[width=0.32\textwidth,height=3.5cm]{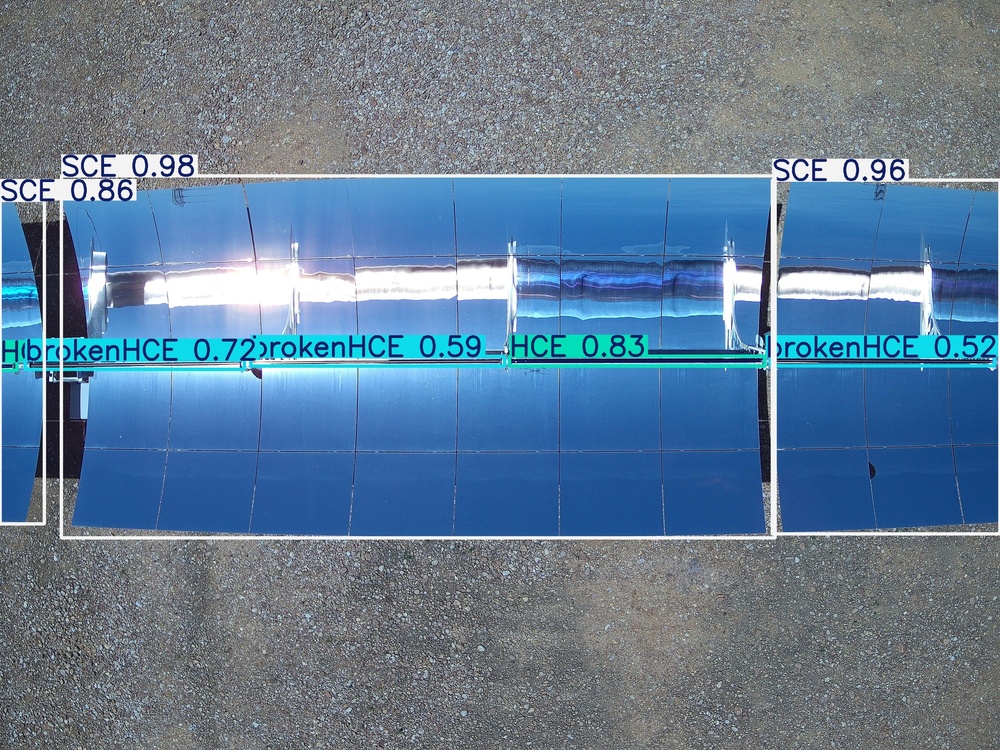}
        \caption{}
    \end{subfigure}
    
    \caption{Qualitative results for object detection on real-world images. The first column shows the input image; the second column displays predictions from YOLOv11x without pretraining on AerialCSP; and the third column shows predictions from the same model pretrained on AerialCSP. Each example illustrates the benefits of domain-specific pretraining. Without pretraining, the model struggles in scenarios involving (a) high brightness, (b) rotated components, and (c) damaged HCEs. In example (c), HCEs without glass coverings can be visually identified through their reflections in nearby mirrors.}
    \label{fig:qualitative_results_broken_set}
\end{figure*}

Table \ref{tab:realds400_detection_results} reinforces the effectiveness of pretraining by showing the results for the case where 400 images were used in the training set. Notably, this is the scenario where non-pretrained models achieve their closest performance to pretrained counterparts. However, these overall results do not account for individual class performance, potentially masking the poor detection capability of non-pretrained models for critical object categories. To address this, Table \ref{tab:realds400_detection_results_broken_things} provides a breakdown specifically for the broken HCE and broken mirror classes, two of the most important categories for CSP plant maintenance. The non-pretrained models struggle significantly with these classes due to two key reasons:
\begin{itemize}
    \item Limited training samples. Broken HCEs and mirrors are less frequent in the dataset.
    \item Small object size. The damaged portions of mirrors are often tiny, making them difficult to detect.
\end{itemize}
As shown in the results, mAP50 scores for non-pretrained models are consistently low across all cases, with yolo11n and yolo11m performing especially poorly on the broken HCE class. In contrast, the pretrained models perform significantly better, achieving results much closer to their mAP50 scores for all classes. This demonstrates that pretraining on AerialCSP enables models to generalize well to novel, hard-to-learn object categories, even when they were not originally present in the dataset. To further support this argument, Figure \ref{fig:qualitative_results_broken_set} presents qualitative results from the yolov11x model trained on 400 real images. The advantages of pretraining on AerialCSP are clearly evident, with the pretrained model demonstrating significantly improved detection performance across challenging scenarios.

\section{Conclusions}
\label{sec:conclusions}

In this work, we introduced AerialCSP, a synthetic dataset designed for aerial inspection of CSP plants, and demonstrated its effectiveness in object detection and segmentation tasks. Our experiments highlight the advantages of pretraining on AerialCSP, significantly improving model performance in real-world failure detection, especially for rare and small defects. These findings reinforce the value of synthetic data to accelerate AI adoption in industrial applications, reduce annotation costs, and improve model generalization. 
The proposed background inpainting technique is adaptable across various scenarios. Thus, future work will explore domain adaptation techniques to further bridge the gap between synthetic and real-world imagery. Furthermore, we plan to extend our experiments to other real-world datasets (for example, using thermal images) to better assess the effectiveness of pretraining in AerialCSP.






\section*{ACKNOWLEDGMENT}

The authors express their gratitude to J.M. Medianero for helping with the initial CAD models, and the IMUS cluster computing center at the University of Seville.


\bibliographystyle{elsarticle-num}
\bibliography{references}




\end{document}